\pdfoutput=1
\documentclass[11pt]{article}

\usepackage[final]{coling}

\usepackage{times}
\usepackage{latexsym}

\usepackage[T1]{fontenc}
\usepackage[utf8]{inputenc}

\usepackage{microtype}

\usepackage{inconsolata}
\usepackage{hyperref}
\usepackage{amssymb}
\usepackage[utf8]{inputenc}
\usepackage{CJK}

\usepackage{graphicx}
\usepackage{booktabs}
\usepackage[symbol]{footmisc}

\usepackage{graphicx}
\usepackage{booktabs}
\usepackage[symbol]{footmisc}

\usepackage{url}            % simple URL typesetting
\usepackage{booktabs}       % professional-quality tables
\usepackage{amsfonts}       % blackboard math symbols
\usepackage{nicefrac}       % compact symbols for 1/2, etc.
\usepackage{xcolor}         % colors
\usepackage{graphicx}
\usepackage{amsmath}
\usepackage{cleveref}  % must be loaded after amsmath
\usepackage{colortbl}
\usepackage{tablefootnote}
\usepackage{algorithm}
\usepackage{algorithmic}
\usepackage{multirow}
\usepackage{colortbl}
\usepackage{caption}
\usepackage{wrapfig}
\usepackage{comment}
\usepackage[para,online,flushleft]{threeparttable}
\usepackage[export]{adjustbox}

\definecolor{light-gray}{gray}{0.8}

\title{ProSparse: Introducing and Enhancing Intrinsic Activation Sparsity\\within Large Language Models}

\author{Chenyang Song$^{1}$, Xu Han$^{1*}$, Zhengyan Zhang$^{1}$, Shengding Hu$^{1}$, Xiyu Shi$^{2}$\\
\textbf{Kuai Li$^{3}$}\textbf{, Chen Chen$^{3}$}\textbf{, Zhiyuan Liu$^{1}$\thanks{ The corresponding authors of this paper: Xu Han (han-xu@tsinghua.edu.cn) and Zhiyuan Liu (liuzy@tsinghua.edu.cn).}}\textbf{, Guangli Li$^{2}$}\textbf{, Tao Yang$^{3}$}\textbf{, Maosong Sun}$^{1}$\\
\textsuperscript{1} Dept. of Comp. Sci. \& Tech., Institute for AI, Tsinghua University, Beijing, China\\
\textsuperscript{2} SKLP, Institute of Computing Technology, Chinese Academy of Sciences, China\\
\textsuperscript{3} Tencent Machine Learning Platform, China\\
\texttt{scy22@mails.tsinghua.edu.cn, \{han-xu,liuzy\}@tsinghua.edu.cn}
}

\begin{document}
\maketitle
\begin{abstract}

Activation sparsity refers to the existence of considerable weakly-contributed elements among activation outputs, serving as a promising paradigm for accelerating model inference.
Nevertheless, most large language models (LLMs) adopt activation functions without intrinsic activation sparsity (e.g., GELU and Swish).
Some recent efforts have explored introducing ReLU or its variants as the substitutive activation function to pursue activation sparsity and acceleration, but few can simultaneously obtain high activation sparsity and comparable model performance.
This paper introduces a simple and effective method named ``ProSparse'' to sparsify LLMs while achieving both targets.
Specifically, after introducing ReLU activation, ProSparse adopts progressive sparsity regularization with a factor smoothly increasing for multiple stages. This can enhance activation sparsity and mitigate performance degradation by avoiding radical shifts in activation distributions.
With ProSparse, we obtain high sparsity of 89.32\% for LLaMA2-7B, 88.80\% for LLaMA2-13B, and 87.89\% for end-size MiniCPM-1B, respectively, with comparable performance to their original Swish-activated versions. These present the most sparsely activated models among open-source LLaMA versions and competitive end-size models.
Inference acceleration experiments further demonstrate the significant practical acceleration potential of LLMs with higher activation sparsity, obtaining up to 4.52$\times$ inference speedup.

\end{abstract}

\section{Introduction}

Recent years have witnessed significant breakthroughs made by large language models (LLMs) with commendable performance across a wide range of NLP tasks~\cite{brown2020language,wei2021finetuned,ouyang2022training,chatgpt,achiam2023gpt}.
Nevertheless, the formidable computational costs required by the deployment and inference of LLMs pose a considerable challenge~\cite{aminabadi2022deepspeed,pope2023efficiently}. 
The utilization of activation sparsity is one of the most promising techniques to enhance inference efficiency~\cite{liu2023deja,song2023powerinfer}, by discarding the redundant computation associated with the elements among LLM activation outputs that contribute weakly to the final outputs. 

\begin{figure*}[ht]
    \centering
    \includegraphics[width=0.9\linewidth]{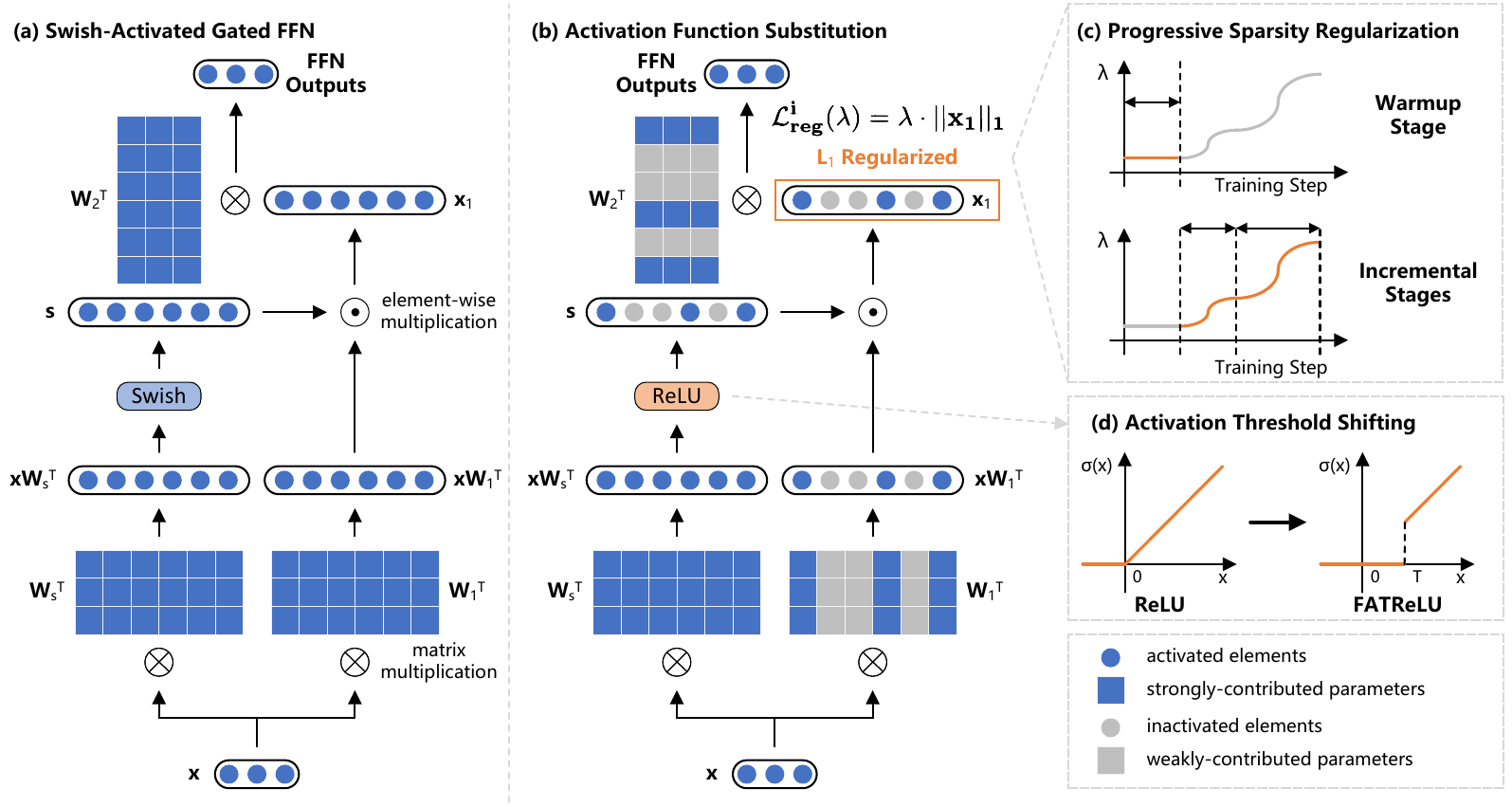}
    \caption{The overall architecture of ProSparse, which includes three steps: activation function substitution, progressive sparsity regularization, and activation threshold shifting.}
    \label{fig:prosparse}
    \vspace{-1em}
\end{figure*}

The adoption of ReLU, which naturally outputs zero elements, as the activation function is a straightforward method to achieve intrinsic activation sparsity in early LLMs~\cite{raffel2020exploring,zhang2022opt}.
However, recent LLMs predominantly favor non-ReLU activation functions, such as GELU and Swish~\cite{touvron2023llama,chowdhery2023palm,almazrouei2023falcon}. Although these non-ReLU LLMs may also display activation sparsity~\cite{zhang2024relu}, such sparsity is manually imposed by searching adaptive activation thresholds (i.e., non-intrinsic), which can potentially lose minor neuron outputs and degrade performance.
To pursue the sparsity-based inference acceleration, the task of ReLUfication is proposed, aiming to introduce ReLU-based intrinsic activation sparsity into non-ReLU LLMs.
Preliminary methods~\cite{zhang2022moefication,zhang2024relu} directly substitute the activation functions with ReLU.
As such substitution cannot overcome the inherent limitation imposed by the original dense activation distribution, inserted and shifted ReLU functions~\cite{mirzadeh2023relu} are introduced to enforce higher sparsity through radically shifting the activation distribution.
However, existing efforts fail to achieve satisfactory sparsity and risk performance degradation.

In this paper, we propose a simple and effective ReLUfication method named ``\textbf{ProSparse}'' to help non-ReLU LLMs obtain high activation sparsity without performance degradation. ProSparse includes three steps shown in Figure~\ref{fig:prosparse}: activation function substitution, progressive sparsity regularization, and activation threshold shifting.
The first step is to replace the activation function with ReLU and then apply continual training. As discussed above, this can hardly achieve satisfactory sparsity.
Therefore, in the second step, we apply sparsity regularization~\cite{ma2019transformed} to the intermediate activation outputs of the feed-forward networks (FFNs) within LLMs to seek higher sparsity. Considering the potential performance risks of forcing the fixed regularization factor~\cite{ma2019transformed,li2020efficacy}, we progressively increase the factor in multiple stages, including one flat warmup stage and multiple incremental stages along gentle sine curves.
Such progressive regularization can provide more time for adaption to increasing regularization and avoid a radical shift in activation distribution, thereby mitigating performance degradation.
The final step adopts FATReLU~\cite{kurtz2020inducing}, shifting the ReLU activation threshold to a positive value. This prunes less influential neurons to improve sparsity.

% Concretely, the factor is set to a low constant value for the warmup stage. Next, during each subsequent stage, the factor undergoes a gradual increase along a gentle sine curve.

\setcounter{footnote}{0}

In experiments, we apply ProSparse to the ReLUfication of LLaMA2~\cite{touvron2023llama2} and end-size MiniCPM~\cite{hu2024minicpm}. Activation sparsity of 89.32\%, 88.80\%, and 87.89\% are successfully achieved for LLaMA2-7B, LLaMA2-13B, and MiniCPM-1B, respectively, with performance comparable to their original Swish-activated versions on various LLM benchmarks. 
Furthermore, we demonstrate the practical inference acceleration of higher activation sparsity, by respectively applying an approximate algorithm and an accurate algorithm to the inference of models with different sparsity.
For the approximate one, we use PowerInfer~\cite{song2023powerinfer}, which achieves state-of-the-art speedup ratios tailored for sparsely activated LLMs but risks inaccurate inference due to the mistakes of activation predictors. 
For the accurate one, we implement and release two GPU operators that leverage the input-side and output-side sparsity during the computation of ReLU-activated FFNs\footnote{Source codes for these two operators are available at \url{https://github.com/Raincleared-Song/sparse_gpu_operator}.}.

The experimental results demonstrate that models with higher sparsity can achieve more significant inference acceleration with both approximate and accurate algorithms (e.g., up to 4.52$\times$ speedup with PowerInfer).
Moreover, comprehensive analyses are conducted to figure out the quantitative relationship between the activation sparsity and the regularization factor, making the activation sparsity obtained by ProSparse more controllable. We also discuss the rationality of progressive $L_1$ regularization, empirical methods of performing supervised fine-tuning (SFT) on sparsely activated models, and the sparsity distribution among distinct datasets or layers.

In summary, we make the following contributions in this paper: 
(1) We propose ProSparse, an effective ReLUfication method that 
converts non-ReLU LLMs into much sparser ReLU-activated LLMs without performance degradation. 
(2) Sparsely activated versions of LLaMA2-7B, LLaMA2-13B, and MiniCPM-1B comparable to their original Swish-activated versions in performance are both obtained and available\footnote{Models are respectively available at \url{https://huggingface.co/SparseLLM/prosparse-llama-2-7b}, \url{https://huggingface.co/SparseLLM/prosparse-llama-2-13b}, and \url{https://huggingface.co/SparseLLM/ProSparse-MiniCPM-1B-sft}.}.
(3) We demonstrate the practical inference acceleration effect of higher activation sparsity that ProSparse can reach. Valuable observations and analyses are also conducted.

\section{Preliminaries and Related Works}

Here we discuss how to improve LLM inference efficiency. Refer to existing surveys~\cite {zhao2023survey} for works about LLMs and Appendix~\ref{sec:more-related-work} for works about $L_1$ regularization.

% bommasani2021opportunities

\paragraph{Inference Acceleration for LLMs}
Efficiency has long been a crucial topic in various AI applications~\cite{chen2023future}. The sustainable increase in LLM scales brings the exponential growth of inference computations, making the deployment of LLMs a formidable challenge~\cite{kaplan2020scaling,liu2023deja}. 
To reduce the computational costs required by LLM inference, various model compression or decoding acceleration methods have been proposed, such as quantization~\cite{jacob2018quantization,nagel2019data,zhao2019improving,bai2022towards,xiao2023smoothquant,yao2023comprehensive}, pruning~\cite{hoefler2021sparsity,ma2023llm,sun2023simple,frantar2023sparsegpt,xia2023sheared}, distillation~\cite{tang2019distilling,touvron2021training,gu2023knowledge,hsieh2023distilling}, and efficient sampling methods~\cite{leviathan2023fast,wang2023tabi,chen2023accelerating,miao2023specinfer}. While these works have proved effective for inference acceleration and other scenarios (e.g., secure federated learning~\cite{ding2023secure}), none of these methods leverages the intrinsic mechanisms within LLMs.

% han2015deep,
% han2015deep,han2015learning,molchanov2016pruning,
% hinton2015distilling,

\paragraph{Activation Sparsity}
Recent works~\cite{li2022lazy,liu2023deja,song2023powerinfer} have noticed the intrinsic activation sparsity within some LLMs and its potential in inference acceleration.
Activation sparsity refers to the phenomenon where considerable zero or negligible elements in activation outputs, corresponding to certain model parameters (i.e., neurons), have a weak impact on LLM outputs given a specific input.
These weakly-contributed parameters are regarded as inactivated and can thus be skipped during inference to save computational resources.
Notably, \textbf{the utilization of activation sparsity is orthogonal to model compression and efficient sampling, and these approaches can be readily combined}. Another fact worth attention is \textbf{the fundamental difference between activation sparsity and pruning}, see Appendix~\ref{sec:more-related-work}.

\paragraph{ReLUfication}
Activation sparsity naturally exists in ReLU-activated architecture~\cite{li2022lazy}, from LLMs~\cite{raffel2020exploring,zhang2022opt} to vision models~\cite{dosovitskiy2020image}.
However, recent LLMs such as Falcon~\cite{almazrouei2023falcon} and LLaMA~\cite{touvron2023llama2} prevalently adopt non-ReLU activation functions such as GELU~\cite{hendrycks2016gaussian} and Swish~\cite{elfwing2018sigmoid} without intrinsic activation sparsity.
Therefore, to leverage the merits of activation sparsity without training a ReLU-activated LLM from scratch, many works conduct ReLUfication, introducing sparse ReLU-based activations into non-ReLU LLMs. \citet{zhang2022moefication} converts a GELU-activated BERT~\cite{devlin2018bert} into a ReLU-activated version through activation function substitution and additional training. ReluLLaMA and ReluFalcon apply a similar paradigm to Falcon and LLaMA, respectively~\cite{zhang2024relu}. 
Since activation substitution cannot reach satisfactory sparsity, mainly due to the unhandled limitation of the original dense activation distribution, the inserted and shifted ReLU activation functions are introduced~\cite{mirzadeh2023relu}, conducting a radical shift in activation distribution. Although these operations are claimed to achieve sparsity of nearly 95\%, we cannot replicate the results in our experiments (see the 3rd paragraph of Section~\ref{sec:discussion}) and the sparsity is still limited. By contrast, ProSparse is a ReLUfication method designed to achieve high sparsity and mitigate performance degradation concurrently.

% As discussed above, we can clearly recognize the promise of activation sparsity and also observe the key challenge of leveraging ReLUfication to achieve activation sparsity: how to achieve high sparsity and mitigate performance degradation concurrently. These are what ProSparse is designed for.

% To this end, this paper introduces ProSparse, a ReLUfication method that can obtain high ReLU-based activation sparsity for non-ReLU LLMs with comparable performance to the original non-ReLU models.

\section{Methods}

\subsection{Definitions and Notations}

For the convenience of subsequent demonstrations, here we define activation sparsity in detail. Since the activation function mainly exists in the FFNs within LLMs, we first discuss the computation process of FFNs. Given the hidden dimension $d_{model}$ and the intermediate dimension $d_{ff}$, the computation process of a gated FFN (i.e., the most widely adopted FFN architecture in recent LLMs~\cite{dauphin2017language,shazeer2020glu}) can be formalized as:
\begin{equation}
    \begin{aligned}
    \label{eq:process-ffn}
    \mathbf{s} = \sigma(\mathbf{x} \mathbf{W}_s^T),  \quad &\mathbf{x}_1 = \mathbf{s} \odot (\mathbf{x} \mathbf{W}_1^T),\\
    \text{FFN}(\mathbf{x}) &= \mathbf{x}_1  \mathbf{W}_2^T,
    \end{aligned}
\end{equation}
% \noindent 
where $\mathbf{x}\in\mathbb{R}^{d_{model}}$, $\mathbf{s}, \mathbf{x}_1\in\mathbb{R}^{d_{ff}}$, $\sigma$, and $\odot$ denote the input hidden states, the gating scores, the intermediate outputs, the activation function, and the element-wise multiplication respectively. $\mathbf{W}_s,\mathbf{W}_1\in\mathbb{R}^{d_{ff} \times d_{model}}$ and $\mathbf{W}_2\in\mathbb{R}^{d_{model} \times d_{ff}}$ are learnable weights.

We define the \textbf{activation sparsity} (hereinafter abbreviated as \textbf{sparsity}) as the ratio of zero elements (i.e., inactivated elements) in $\mathbf{x}_1$ for a specific input $\mathbf{x}$. The sparsity of an LLM is evaluated using the \textbf{average sparsity}, defined as the average value of sparsity across all layers in an LLM on a sufficiently large amount of input data.

% When ReLU is adopted for activation, namely $\sigma(x)=\max(x,0)$
% Since the sparsity varies in different layers for different inputs,

In this paper, we focus on the task of ReLUfication, namely converting an LLM using a non-ReLU activation function $\sigma$ (e.g., GELU or Swish) into a ReLU-activated one, while making the average sparsity as high as possible and mitigating performance degradation.

\subsection{ProSparse}

We propose ProSparse to achieve the above targets. Three steps are carefully designed to introduce and enhance the intrinsic activation sparsity for a non-ReLU LLM: (1) activation function substitution; (2) progressive sparsity regularization; (3) activation threshold shifting.

\paragraph{Activation Function Substitution} \label{sec:relu-substitution}

For lack of attention to activation sparsity, a majority of recent mainstream LLMs adopt non-ReLU activation functions such as GELU and Swish that output few zero elements (i.e., low activation sparsity according to the above definition). Therefore, the first step of ProSparse is to introduce intrinsic sparsity through activation function substitution, which replaces the FFN activation function $\sigma$ with ReLU, namely $\sigma(x)=\max(x, 0)$, followed by continual training. This can make the ratio of zero activation elements significantly larger and preliminarily adapt the LLM to new ReLU activation.

\paragraph{Progressive Sparsity Regularization} \label{sec:pro-reg}

Nevertheless, activation function substitution by nature does not change the activation distribution, which will potentially limit the sparsity to relatively low values. To push for higher sparsity, a typical method is $L_1$ sparsity regularization~\cite{li2022lazy}, which introduces the $L_1$ regularization loss as an extra training target. Given the intermediate output $\mathbf{x}_1$ of the $i$-th FFN layer in an LLM, the regularization loss is defined as:
\begin{equation}
    \label{eq:loss-reg}
    \mathcal{L}_{reg}^i(\lambda) = \lambda\cdot||\mathbf{x}_1||_1,
\end{equation}
% \noindent 
where $||\cdot||_1$ is the $L_1$ norm operator and $\lambda$ is the regularization factor. For an LLM with $K$ FFN layers, the total regularization loss is summed from the losses of all layers, namely 
$\mathcal{L}_{reg}(\lambda)=\sum_{i=1}^K \mathcal{L}_{reg}^i(\lambda)$. The overall optimization target is $\mathcal{L}_{lm}+\mathcal{L}_{reg}(\lambda)$, where $\mathcal{L}_{lm}$ is the vanilla language modeling loss.

Considering the potential performance degradation due to fixed regularization factors~\cite{georgiadis2019accelerating,kurtz2020inducing,li2022lazy}, we propose the progressive sparsity regularization, where the factor $\lambda$ is carefully scheduled to gently increase in multiple stages. Refer to Appendix~\ref{sec:algorithm-pro} for more details.

Concretely, for the warmup stage, we set $\lambda$ to a constant value, which is relatively small to prevent radical activation distribution shifts and introduce higher preliminary sparsity. Next, for each of the remaining stages (called incremental stages), $\lambda$ is scheduled to increase along a smooth sine curve from a trough value to a peak value. Inspired by the cosine annealing scheduler for learning rates~\cite{loshchilov2016sgdr}, we choose the sine function owing to its special trend, as the small derivatives near its troughs and peaks can make $\lambda$ not increase radically around these two points. This provides the LLMs with more time to adapt the activation distributions to the newly increased $L_1$ regularization. Notably, each stage is accompanied by certain steps of training. The step numbers and peak $\lambda$ values are chosen according to the target sparsity and stability.

\paragraph{Activation Threshold Shifting} \label{sec:thres-adj}

As demonstrated by recent works, there exist considerable amounts of non-zero low elements in the activation outputs, which have little influence on final results and thus can be pruned for higher sparsity~\cite{zhang2024relu}. Therefore, we convert the ReLU into FATReLU~\cite{kurtz2020inducing} by shifting the activation threshold, i.e., 
\begin{equation}
    \sigma(x)=
    \begin{cases}
    x \quad \mathrm{when}\ x \geq t, \\
    0 \quad \mathrm{otherwise},
    \end{cases}
    \label{eq:fat-relu}
\end{equation}
% \noindent 
where $t>0$ is a positive threshold. As long as $t$ is properly chosen (see Appendix~\ref{sec:different-thres}), FATReLU can increase sparsity with negligible losses~\cite{zhang2024relu}.

\subsection{Practical Inference Acceleration}

To go beyond theoretical analyses based on FLOPS (Floating Point Operations Per Second)~\cite{mirzadeh2023relu} and establish the practical value of ProSparse, we discuss how to realize inference acceleration with sparsely activated LLMs on real hardware and how to evaluate the practical acceleration effects.
We consider two categories of acceleration algorithms based on activation sparsity: approximate algorithms and accurate algorithms.

\paragraph{Approximate Acceleration Algorithms}

Utilizing activation sparsity, recent approximate acceleration algorithms predominantly rely on activation predictors, typically small neural networks, to forecast the activation distributions indicated by the sparse intermediate outputs $\mathbf{x}_1$ given a specific input $\mathbf{x}$~\cite{liu2023deja,song2023powerinfer}. In this way, they can make wiser hardware allocation or computation policies to avoid resource waste on weakly-contributed parameters. However, their efficiency and accuracy largely depend on the predictors' performance, and invalid predictions can cause suboptimal hardware allocation or even inference inaccuracy. Therefore, to gain more practical acceleration effects from approximate algorithms, both high activation sparsity and predictability are indispensable.

To this end, we focus on three metrics for acceleration analysis: the activation recall, the predicted sparsity, and the inference speed. The former two metrics evaluate the performance of activation predictors as well as the activation predictability of a sparse LLM~\cite{zhang2024relu}. For inference speed, we adopt PowerInfer~\cite{song2023powerinfer}, a state-of-the-art approximate algorithm to measure practical speedup ratios. Refer to Appendix~\ref{sec:approximate-alg} for more related introductions and the detailed approach to calculating these metrics.

% (hereinafter abbreviated as recall)

\paragraph{Accurate Acceleration Algorithms}  \label{sec:accurate-alg}

\begin{table*}[ht]
    \centering
    \scriptsize
    \setlength{\tabcolsep}{0.8mm}
    \begin{tabular}{lccccccc>{\columncolor{light-gray}}c>{\columncolor{light-gray}}c}
        \toprule
        \multirow{2}{*}{Setting} & Code & Commonsense & Reading & \multirow{2}{*}{GSM8K} & \multirow{2}{*}{MMLU} & \multirow{2}{*}{BBH} & \multirow{2}{*}{AGI Eval} & Average & Average \\
         & Generation & Reasoning & Comprehension & & & & & Performance & Sparsity \\
        \cmidrule(lr){1-1}\cmidrule(lr){2-8}\cmidrule(lr){9-10}
        LLaMA2-7B  & 16.37 & 69.59 & 61.87 & 12.96 & 44.45 & 32.96 & 27.53 & 37.96 & - \\
        ReluLLaMA-7B & 15.85 & 69.64 & 70.54 &  5.84 & 38.64 & 35.07 & 27.73 & 37.62 & 66.98 \\
        % Vanilla ReLU-7B & 21.31 & 70.73 & 73.22 & 11.22 & 49.22 & 36.11 & 28.01 & 41.40 & 66.04 \\
        % Shifted ReLU-7B & 20.50 & 70.09 & 73.17 & 13.87 & 48.54 & 35.20 & 27.94 & 41.33 & 69.59 \\
        % Fixed $L_1$-7B & 18.85 & 66.01 & 55.39 & 2.27 & 32.28 & 31.40 & 26.48 & 33.24 & 91.46 \\
        \textbf{ProSparse-7B$^*$} & 19.47 & 66.29 & 63.33 & 12.74 & 45.21 & 33.59 & 27.55 & 38.31 & 88.11 \\
        \textbf{ProSparse-7B} & 19.42 & 66.27 & 63.50 & 12.13 & 45.48 & 34.99 & 27.46 & \textbf{38.46} & \textbf{89.32} \\
        \cmidrule(lr){1-1}\cmidrule(lr){2-8}\cmidrule(lr){9-10}
        LLaMA2-13B & 20.19 & 72.58 & 71.55 & 22.21 & 54.69 & 37.89 & 29.33 & 44.06 & - \\
        ReluLLaMA-13B & 20.19 & 70.44 & 73.29 & 18.50 & 50.58 & 37.97 & 28.22 & 42.74 & 71.56 \\
        \textbf{ProSparse-13B$^*$} & 29.03 & 69.75 & 67.54 & 25.40 & 54.78 & 40.20 & 28.76 & \textbf{45.07} & 87.97 \\
        \textbf{ProSparse-13B} & 28.42 & 69.76 & 66.91 & 26.31 & 54.35 & 39.90 & 28.67 & 44.90 & \textbf{88.80} \\
        \cmidrule(lr){1-1}\cmidrule(lr){2-8}\cmidrule(lr){9-10}
        % 41.46 & 32.24 & 74.1 & 48.87 & 67.27 & 59.12 & 69.0 & 69.3 & 58.18 & 55.23
        MiniCPM-1B & 36.85 & 63.67 & 60.90 & 35.48 & 50.44 & 35.03 & 28.71 & 44.44 & - \\
        % 48.78 & 33.98 & 73.99 & 47.54 & 68.46 & 59.75 & 73.0 & 74.92 & 53.74 & 53.41
        \textbf{ProSparse-1B$^*$} & 41.38 & 64.55 & 60.69 & 34.72 & 49.36 & 34.04 & 28.27 & \textbf{44.72} & 86.25 \\
        % FAT 0.01
        % 40.26 & 64.47 & 60.51 & 33.06 & 49.27 & 34.87 & 28.23 & 44.38
        % 48.17 & 32.34 & 74.05 & 47.59 & 68.51 & 60.22 & 72.0 & 74.89 & 53.5 & 53.14
        % FAT 0.03
        % 50.61 & 33.47 & 74.16 & 47.54 & 68.4 & 59.75 & 72.0 & 75.08 & 53.74 & 53.36
        \textbf{ProSparse-1B} & 42.04 & 64.37 & 60.73 & 34.57 & 49.51 & 34.08 & 27.77 & \textbf{44.72} & \textbf{87.89} \\
        \bottomrule
    \end{tabular}
    \caption{The overall experimental results with the comparison of activation sparsity (\%) and downstream performance (\%). ``LLaMA2'' and ``MiniCPM'' refer to the original Swish-activated LLaMA2~\cite{touvron2023llama2} and MiniCPM~\cite{hu2024minicpm} versions respectively. ``ProSparse-7B$^*$'', ``ProSparse-13B$^*$'', and ``ProSparse-1B$^*$'' denote the ProSparse versions \textbf{without activation threshold shifting}.}
    \label{tab:main-results}
    \vspace{-1em}
\end{table*}

Targeting acceleration without potential inference inaccuracies, we implement two hardware-efficient sparse GPU operators with system-level optimizations, such as operator fusion, coalesced memory access, and vectorization, thereby exploiting input-side and output-side sparsity in Equation~\ref{eq:process-ffn}.

Concretely, we reorganize a ReLU-activated gated FFN into three major steps and our two operators are responsible for the step (2) and (3) respectively:
(1) A dense matrix-vector multiplication operator $ \mathbf{x} \mathbf{W}_s^T $ directly supported by vendor libraries such as cuBLAS;
(2) A fused operator of ReLU and $\mathbf{s} \odot (\mathbf{x} \mathbf{W}_1^T)$ with output-side sparsity;
(3) A sparse matrix-vector multiplication operator $\mathbf{x}_1 \mathbf{W}_2^T$ with input-side sparsity. We adopt the single-step speedup ratios of steps (2) and (3) with these two operators respectively to reflect the practical accurate acceleration potential of sparse LLMs. Refer to Appendix~\ref{sec:impl-operator} for implementation details.

\section{Experiments} \label{sec:experiments}

\subsection{Experimental Settings}

Our training data consists of both language modeling datasets and instruction tuning datasets. For evaluation, we adopt comprehensive benchmarks covering code generation, commonsense reasoning, reading comprehension, and 4 other popular tasks. Refer to Appendix~\ref{sec:dataset-detail} for more details.

\subsection{Overall Results} \label{sec:main-results}

We apply ProSparse to Swish-activated LLaMA2-7B, LLaMA2-13B~\cite{touvron2023llama2}, and MiniCPM-1B~\cite{hu2024minicpm}. The obtained sparsely activated models are then compared with their original Swish-activated versions. For comprehensiveness, we also consider ReluLLaMA\footnote{\url{https://huggingface.co/SparseLLM/ReluLLaMA-7B}}, the only open-source ReLU-based LLMs fine-tuned from LLaMA2. All the average sparsity values are computed on the same mixed dataset sampled from training datasets. For more hyper-parameters, see Appendix~\ref{sec:hyper-param} and~\ref{sec:different-thres}.

% The performance can potentially be raised through smoother trends of the regularization factor increase at the cost of more training tokens (see the 3rd paragraph of Section~\ref{sec:discussion}).

% \footnote{\url{https://huggingface.co/meta-llama/Llama-2-7b}}
% \footnote{\url{https://huggingface.co/openbmb/MiniCPM-1B-sft-bf16}}

Results are shown in Table~\ref{tab:main-results} (see Appendix~\ref{sec:performance-component} for performance on each independent benchmark). We can draw three conclusions:
% From the average sparsity and performance scores,

(1) \textit{Effectiveness}: ProSparse simultaneously achieves high sparsity and comparable downstream performance for all the three Swish-activated models considered. The activation sparsity obtained by ProSparse is significantly higher than ReluLLaMA, reaching the state-of-the-art level among all the open-source LLaMA versions and competitive end-size models.

(2) \textit{Scale Generalizability}: The effectiveness of ProSparse consistently holds under three model scales. The promising results on the end-size model (i.e., MiniCPM-1B) reveal the potential of ProSparse as well as activation sparsity on end-user devices, where the inference efficiency of LLMs is significantly emphasized.

(3) \textit{Effect of Activation Threshold Shifting}: Based on the results without activation threshold shifting (i.e., those with the ``$^*$'' marker), we can demonstrate the effectiveness of this technique in improving the sparsity without compromising performance. Notably, the threshold $t$ must be carefully chosen, see Appendix~\ref{sec:different-thres}.

\subsection{Acceleration Effect of Sparsity}

\begin{table*}[t]
    \centering
    \scriptsize
    \setlength{\tabcolsep}{1.5mm}
    \begin{threeparttable}
    \begin{tabular}{lcccc>{\columncolor{light-gray}}cc>{\columncolor{light-gray}}cc>{\columncolor{light-gray}}c}
        \toprule
        & & \multicolumn{4}{c}{Approximate Acceleration} & \multicolumn{4}{c}{Accurate Acceleration} \\
        \cmidrule(lr){3-6}\cmidrule(lr){7-10}
        \multirow{2}{*}{Setting} & Average & Activation & Predicted & Inference & Speedup & Step (2) & Speedup & Step (3) & Speedup \\
        & Sparsity & Recall & Sparsity & Speed & to Dense & Time ($\downarrow$) & to Dense & Time ($\downarrow$) & to Dense \\
        \cmidrule(lr){1-1}\cmidrule(lr){2-2}\cmidrule(lr){3-6}\cmidrule(lr){7-10}
        Dense-7B & - & - & - & 3.67 & 1.00 & 90.55 & 1.00 & 82.92 & 1.00 \\
        ReluLLaMA-7B  & 66.98 & 90.89 & 58.95 & 11.37 & 3.10 & 67.12 & 1.35 & 63.00 & 1.32 \\
        % Vanilla ReLU-7B & 66.04 & 87.72 & 72.57 & 12.04 & 1.06 & 67.85 & 1.33 & 63.28 & 1.31 \\
        % Shifted-7B $b=0.3$ & 70.34 &  &  &  &  &  \\
        % Fixed $L_1$-7B & 91.46 & 94.51 & 82.85 & 19.62 & 1.73 & 40.99 & 2.21 & 54.19 & 1.53 \\
        \textbf{ProSparse-7B$^*$}  & 88.11 & \textbf{93.46} & 75.24 & \textbf{16.30} & \textbf{4.44} & 46.66 & 1.94 & 55.56 & 1.49 \\
        \textbf{ProSparse-7B} & \textbf{89.32} & 92.34 & \textbf{78.75} & - & - & \textbf{45.38} & \textbf{2.00} & \textbf{55.05} & \textbf{1.51} \\
        \cmidrule(lr){1-1}\cmidrule(lr){2-2}\cmidrule(lr){3-6}\cmidrule(lr){7-10}
        Dense-13B & - & - & - & 1.92 & 1.00 & 131.36 & 1.00 & 113.68 & 1.00 \\
        ReluLLaMA-13B & 71.56 & 86.41 & 71.93 &  6.59 & 3.43 & 69.92 & 1.88 & 75.47 & 1.51 \\
        \textbf{ProSparse-13B$^*$}  & 87.97 & 91.02 & 77.93 & \textbf{8.67} & \textbf{4.52} & 55.29 & 2.38 & 67.50 & 1.68 \\
        \textbf{ProSparse-13B} & \textbf{88.80} & \textbf{91.11} & \textbf{78.28} & - & - & \textbf{53.78} & \textbf{2.44} & \textbf{66.73} & \textbf{1.70} \\
        \bottomrule
    \end{tabular}
    \begin{tablenotes}
        $\dagger$ For ``Dense'' settings, the ``Inference Speed'' is obtained by llama.cpp, and the time for steps (2) and (3) is measured without sparse GPU operators. For other sparse settings, the ``Inference Speed'' is obtained by PowerInfer, and sparse GPU operators are applied. ProSparse settings with activation threshold shifting and the MiniCPM architecture are not supported by PowerInfer at present.
    \end{tablenotes}
    \captionsetup{skip=2mm}
    \caption{The comparison of activation recalls (\%), predicted sparsity (\%), inference speeds (tokens per second) by llama.cpp (Dense) or PowerInfer (others), and the average wall-clock time (us) without (Dense) or with (others) our sparse GPU operators among LLMs with different sparsity. ``Step (2)'' and ``Step (3)'' correspond to the steps in Section~\ref{sec:accurate-alg}.}
    \label{tab:predict-operator}
    \end{threeparttable}
    \vspace{-2em}
\end{table*}

\paragraph{Approximate Acceleration Algorithm}

In this section, we train the activation predictors for each sparse LLM and compute the recalls, predicted sparsity, and actual inference speeds on PowerInfer~\cite{song2023powerinfer}. As the FFN in each Transformer layer has different activation distributions as well as different predictors, the former two metrics are averaged from the results of all layers. Note that MiniCPM-1B has not been tested since PowerInfer does not support its architecture at present. Refer to Appendix~\ref{sec:train-act-predictor} for training details of predictors.

As demonstrated by the results shown in Table~\ref{tab:predict-operator}, compared with llama.cpp\footnote{\url{https://github.com/ggerganov/llama.cpp}}, an acceleration framework without sparsity utilization, PowerInfer achieves up to 4.52$\times$ speedup, revealing the significant potential of sparsity-based acceleration. Moreover, an increased activation sparsity can considerably improve the activation recall, the predicted sparsity, and the inference speed of PowerInfer. This proves the considerable practical values of even more sparsely activated LLMs in improving the inference speed with predictor-based approximate acceleration algorithms and mitigating the inaccurate inference problem. ProSparse, which reaches a high sparsity without performance degradation, can thus gain the most acceleration effects with PowerInfer.

\vspace{-2mm}

\paragraph{Accurate Acceleration Algorithm}

Furthermore, with LLMs of different sparsity, we measure the average single-step wall-clock time spent by our two sparse GPU operators, which are responsible for step (2) and step (3) in Section~\ref{sec:accurate-alg} respectively. As demonstrated in Table~\ref{tab:predict-operator}, higher activation sparsity can make accurate algorithms based on GPU operators more efficient. Besides, our two sparse GPU operators also display satisfactory speedup ratios up to 2.44$\times$ and 1.70$\times$ respectively with better acceleration effects for larger models. Note that despite the less significant acceleration effects than PowerInfer, our GPU operators are highly pluggable, predictor-free, and immune to potential inference accuracies.

\subsection{Analysis and Discussion} \label{sec:discussion}

\paragraph{\textit{Q1:} What is the effect of $L_1$ regularization and its trend of increase?}
For this question, we consider two regularization-free ReLUfication baselines: vanilla ReLU~\cite{zhang2024relu} and shifted ReLU~\cite{mirzadeh2023relu}. Both only include the substitution of activation functions with ReLU (i.e., $\max(x,0)$) and shifted ReLU (i.e., $\max(x-b, 0)$, where $b$ is a tunable bias) respectively.

\begin{figure}[ht]
    \centering
    \includegraphics[width=0.9\linewidth]{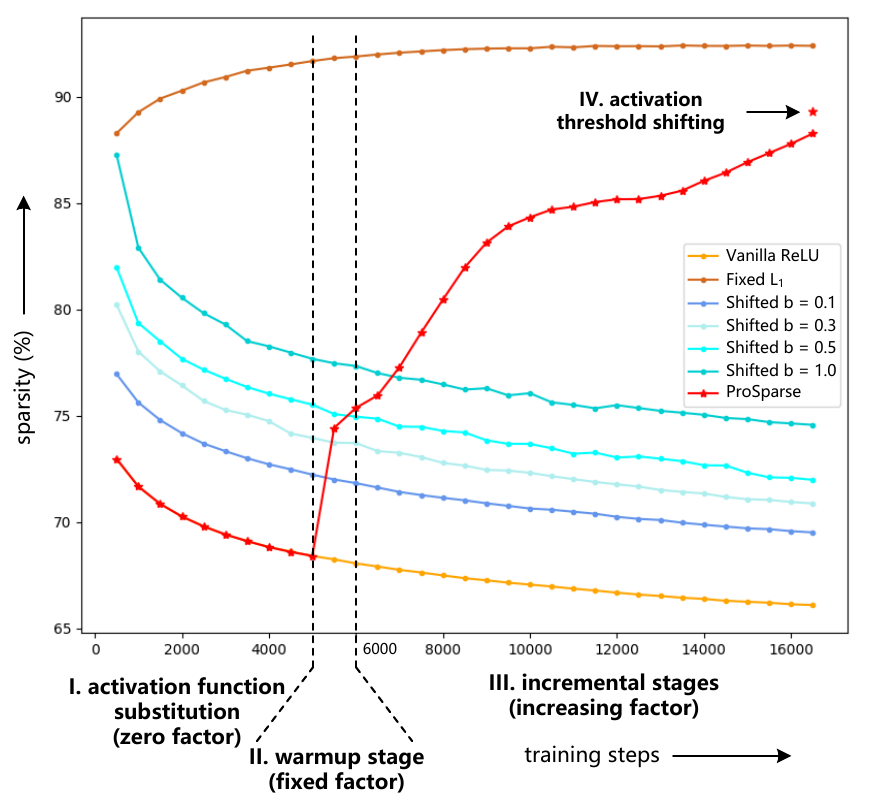}
    \caption{The trend of sparsity (7B models) along the training process. ``Shifted'' denotes Shited ReLU and $b=0.1$ corresponds to the results in Table~\ref{tab:baseline}.}
    \vspace{-0.5em}
    \label{fig:sparsity-curve}
\end{figure}

First, we consider the training dynamics of the above two baselines and ProSparse, as shown in Figure~\ref{fig:sparsity-curve}. The setting ``Fixed $L_1$'' is a reference setting with a constant regularization factor. Clearly, the training stages with increasing sparsity only include those with regularization applied, namely the whole ``Fixed $L_1$'', the warmup stage, and the incremental stages of ProSparse. Therefore, \textbf{among the settings involved, the trend of sparsity is incremental only if non-zero $L_1$ regularization is applied}\footnote{We did not reproduce the flat sparsity trend claimed in the paper of Shifted ReLU~\cite{mirzadeh2023relu}.}. \textbf{Neither vanilla ReLU nor shifted ReLU can push for higher sparsity without regulatization}.

However, concerns may naturally arise about the performance, as the additional $L_1$ loss term can unavoidably influence the optimization of the language modeling target. For this problem, we evaluate the above methods given different numbers of training tokens. Through experiments (see Appendix~\ref{sec:comparison-baseline}), while a performance gap exists between ProSparse and two baselines given limited 34.60B tokens, it obtains comparable performance when sufficient 89.13B tokens are provided and thus the regularization can increase more smoothly, with a final sparsity value close to the limited-token setting. Therefore, \textbf{$L_1$ regularization can reach far higher activation sparsity and maintain comparable performance to regularization-free methods with a sufficiently smooth increase trend of the factor, at the cost of an acceptable rise in training tokens} (i.e., 54.53B, only \textbf{2.73\%} of the 2T tokens used to pre-train LLaMA2~\cite{touvron2023llama2}).
\vspace{-2mm}

% Note that given the same target sparsity, ProSparse can apply a smoother trend of regularization increase with more training tokens (i.e., training steps) available for progressive regularization.

\paragraph{\textit{Q2:} How to reach a target activation sparsity value?}
\begin{figure}[ht]
  \centering
  \includegraphics[width=0.9\linewidth]{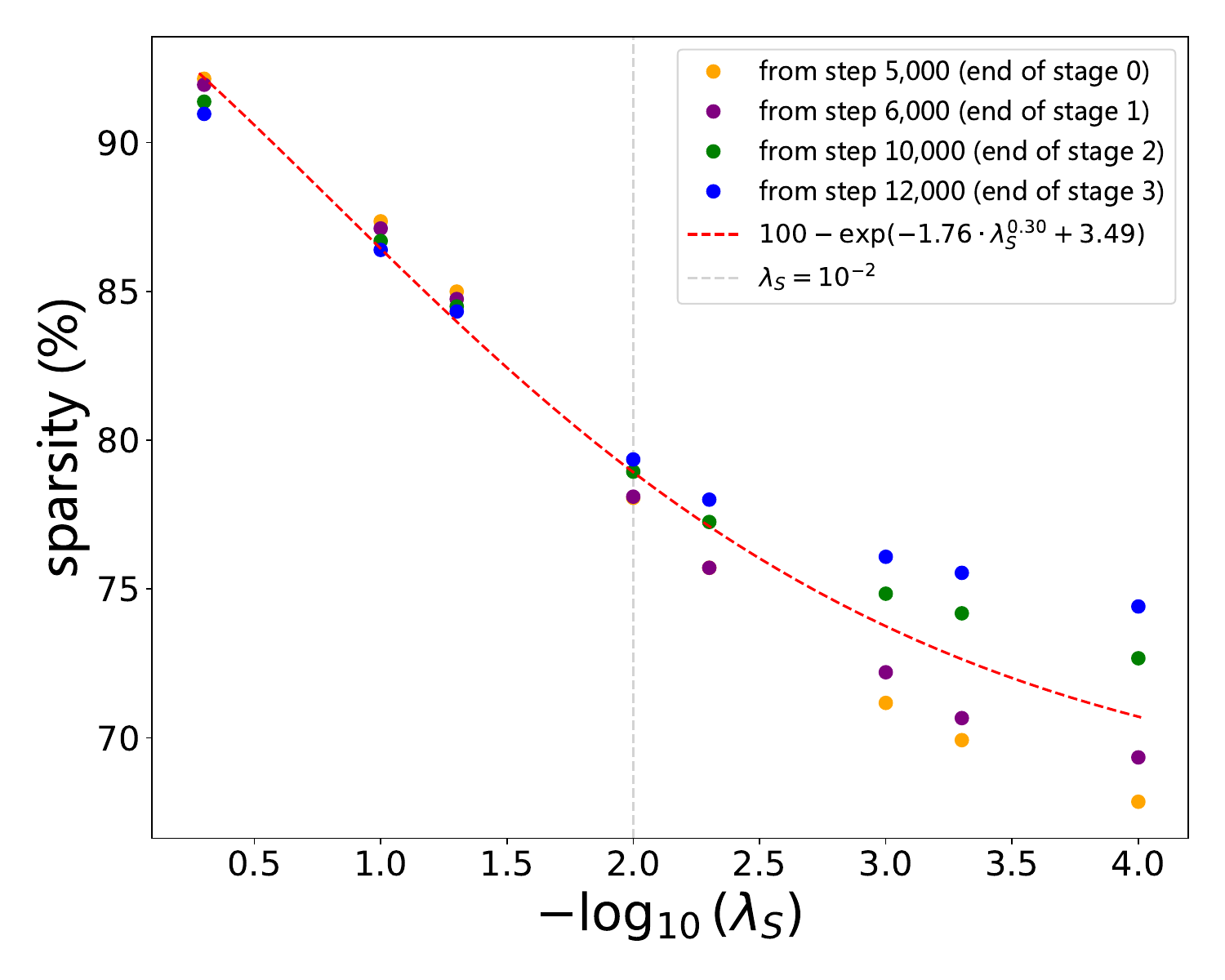}
  \caption{The activation sparsity obtained by applying different final-stage regularization factors $\lambda_S$ to the checkpoints at different training stages (16,500 steps in total) of ProSparse-7B.}
  % \vspace{-1em}
  \label{fig:sparsity-lambda}
\end{figure}

Considering the additional training costs needed to reach higher sparsity, a common requirement lies in how to manipulate ProSparse to reach a desired sparsity given a limited computation budget. The key challenge is the search for suitable regularization factors. To this end, we manage to find the quantitative relationship between the final activation sparsity and the regularization factors to avoid the empirical hyper-parameter search.

Specifically, we select checkpoints at different training stages of ProSparse-7B (see Table~\ref{tab:hyper-param}), apply a constant regularization factor $\lambda_S$, and then resume training for sufficient steps (i.e., no less than 4,000 steps) as the last regularization stage. With the same accumulated training token numbers as ProSparse-7B, we can obtain different activation sparsity by tuning the value of $\lambda_S$. The results shown in Figure~\ref{fig:sparsity-lambda} provide two observations: (1) \textbf{The final activation sparsity is mainly dependent on the last-stage regularization factor $\lambda_S$ when $\lambda_S$ is relatively large} (e.g., $\lambda_S>=10^{-2}$ for ProSparse-7B). (2) \textbf{The activation sparsity shows a negative exponential relationship with $\lambda_S^\alpha$.} For ProSparse-7B, specifically, the sparsity approximates $100-\exp(-1.76\cdot\lambda_S^{0.30}+3.49)$ (i.e., the red fitted curve in Figure~\ref{fig:sparsity-lambda}). In summary, to reach a given relatively high sparsity level (e.g., sparsity larger than 80\%, satisfying $\lambda_S>=10^{-2}$), the only thing needed is to control the regularization factor $\lambda_S$ of the final progressive regularization stage. Therefore, \textbf{given the fixed model size, ProSparse is a highly controllable ReLUfication method in terms of sparsity adjustment}.

\vspace{-1mm}

\paragraph{\textit{Q3:} Is progressive sparsity regularization effective?}

If the activation sparsity mainly depends on the final-stage regularization factor, why should we increase the factor progressively? The answer lies in the performance concern. To substantiate the effectiveness of progressive sparsity regularization, the second step of ProSparse, we conduct ablation studies by making the regularization factor constant throughout the training process after activation function substitution. By setting the factor to $0.1$, we obtain a model with activation sparsity of 88.62\%, slightly lower than ProSparse-7B (89.32\%). However, with the same number of training tokens, this model only has an average performance of 36.34\%, considerably lower than ProSparse-7B (38.46\%). Similarly, for the 13B setting, we obtain a model with the comparable sparsity of 88.96\% and an average performance of 42.85\%, lower than ProSparse-13B (see Appendix~\ref{sec:ablation-progress}). Therefore, \textbf{progressive sparsity regularization is indispensable in mitigating the performance degradation caused by ReLUfication}.

\vspace{-1mm}

\paragraph{\textit{Q4:} How to SFT sparsely activated models?}
It is non-trivial to apply SFT to sparsely activated models obtained by ProSparse. Our practice of training ProSparse-1B can provide some experience: \textbf{SFT can be applied to sparsely activated models obtained by ProSparse with a well-chosen regularization factor, and this factor for SFT is empirically smaller than $\lambda_S$ to accommodate newly injected knowledge and avoid performance degradation}. See Appendix~\ref{sec:sft-sparse} for more details and observations.
\vspace{-2mm}

% 不能没有约束，否则稀疏度会反弹
% 约束不能和预训练一样，因为数据分布变了
% 经验少，我们MiniCPM SFT的约束比预训练小一些

\paragraph{\textit{Q5:} How does the sparsity distribute?}
Another interesting observation is the imbalanced sparsity distributions among distinct datasets and layers. Specifically, \textbf{the activation sparsity of ProSparse models is higher on more formatted instruction tuning datasets and higher layers (i.e., layers closer to outputs)}. More detailed analyses are provided in Appendix~\ref{sec:dataset-dis} and~\ref{sec:layer-dis}.
\vspace{-2mm}

\section{Conclusion}
\vspace{-2mm}

In this work, we propose ProSparse, an effective ReLUfication method for introducing and enhancing intrinsic activation sparsity from non-ReLU LLMs with comparable performance. Extensive experiments demonstrate the effectiveness of ProSparse and its practical values in inference acceleration with various algorithms. Deeper analyses concerned with certain ProSparse techniques, model properties, and SFT issues further substantiate the practicality of ProSparse and provide valuable insights.

% training dynamics, the sparsity control, the rationality of progressive regularization, SFT on ProSparse models, and the sparsity distributions

\section*{Broader Impacts} \label{sec:impact}

This paper presents a simple and effective method, ProSparse, for introducing and enhancing ReLU-based intrinsic activation sparsity into non-ReLU LLMs. There may exist many potential societal consequences of our work, none of which we feel must be specifically highlighted here.

\section*{Limitations} \label{sec:limitations}

Firstly, more comprehensive studies on huge-scale models (e.g., 70B or more) should be included in the future given sufficient computing resources. Moreover, we only focus on the sparsity-based acceleration of step (2) and step (3) of FFN, leaving a considerable ratio of LLM computation unoptimized. Actually, there already exist preliminary works in the sparsification of the attention layers~\cite{shen2023study,wortsman2023replacing}. Methods such as pruning and low-rank decomposition may also be helpful in optimizing the FFN step (1)~\cite{ji2024featurebased}. For future works, we will continue to explore how to introduce and enhance sparsity in the attention layer as well as the acceleration issue of the FFN step (1).

\section*{Acknowledgments}

This work is supported by National Natural Science Foundation of China (No. 62236004, No. 62236011, No. 62302479), China Postdoctoral Science Foundation (2023M733566), and Institute Guo Qiang at Tsinghua University.

\bibliography{custom}

\clearpage

\appendix

\section{Extended Related Works} \label{sec:more-related-work}

\paragraph{$L_1$ Regularization} In statistical learning such as linear regression, $L_1$ regularization has been long adopted as a classical and effective technique for sparsification~\cite{tibshirani1996regression,hastie2009elements}. With the advent of deep learning, researchers also explore paradigms of applying $L_1$ regularization to neural networks. One prominent usage is model pruning~\cite{cheng2017survey}. Specifically, a term of loss calculated as the $L_1$ norm of the sparsification target is added to the optimization target function to prompt sparse weights for faster computation. This has helped acceleration in various conventional neural networks~\cite{han2015learning,zhao2016loss,wen2016learning,scardapane2017group,ma2019transformed,wang2019structured} as well as Transformer-based models~\cite{zhu2021vision,prasetyo2023sparse}. Inspired by these works, some researchers also try to adopt $L_1$ regularization for activation sparsity, mainly in ReLU-activated convolutional networks~\cite{georgiadis2019accelerating,kurtz2020inducing} and Transformer-based architectures~\cite{li2022lazy}.

To the best of our knowledge, ProSparse is the first work using a dynamic $L_1$ regularization factor for prompting activation sparsity in neural networks. By contrast, a majority of the former works adopt fixed factors. For more adaptive control, some of them introduce group regularization~\cite{yuan2006model}, namely using different factors for different parameter groups. Nevertheless, without dynamic factors, these paradigms can cause a substantial shift in activation distribution and thus potentially risk performance degradation. The work most related to ProSparse is IncReg~\cite{wang2019structured}, which introduces incremental regularization factors that change for different parameter groups at each iteration. While they focus on the pruning of convolutional networks, ProSparse handles a distinct scenario of prompting activation sparsity in Transformer-based LLMs and adopts a completely different strategy consisting of a progressively incremental factor.

\paragraph{Difference between Activation Sparsity and Pruning}
Generally, pruning realizes sparsity by removing certain elements (e.g., neurons, weights, or structured blocks) in LLMs. However, the sparsity introduced by pruning is statically limited to model weights and independent of the inputs. High static sparsity is often accompanied by considerable performance degradation compared to the original dense model~\cite{frantar2023sparsegpt,xia2023sheared}. By contrast, activation sparsity is dynamically determined by the input data and thus potentially compromises less model capacity and downstream task performance.
% is the genre most related to activation sparsity. 

\section{Detailed Algorithm for Progressive Sparsity Regularization} \label{sec:algorithm-pro}

The detailed algorithm for scheduling the factor $\lambda$ in progressive sparsity regularization is listed in Algorithm~\ref{al:prosparse-lambda}.

% \begin{wrapfloat}{algorithm}[18]{r}{0.5\textwidth}
\begin{algorithm}[ht]
    \caption{Progressive factor scheduling adopted in ProSparse}
    \label{al:prosparse-lambda}
    \begin{algorithmic}[1]
        \REQUIRE The total number of stages $S\geq 1$.
        \REQUIRE A sequence of peak $\lambda$ values $\{\lambda_i\}_{i=1}^S$, s.t. $0<\lambda_1\leq\lambda_2\leq...\leq\lambda_S$.
        \REQUIRE Accumulated step numbers of each stage $\{T_i\}_{i=1}^S$, s.t. $0<T_1<T_2<...<T_S$.

        % \STATE \texttt{// the first stage}
        % \FOR{$t\leftarrow 1$ to $T_1$}
        %     \STATE $\lambda \leftarrow \lambda_1$, $\lambda_1=0$
        % \ENDFOR
        \STATE \texttt{// warmup stage}
        \FOR{$t\leftarrow 1$ to $T_1$}
            \STATE $\lambda \leftarrow \lambda_1$
            \STATE update model by loss $\mathcal{L}_{lm}+\mathcal{L}_{reg}(\lambda)$
        \ENDFOR
        \STATE \texttt{// incremental stages}
        \FOR{$i\leftarrow 2$ to $S$}
            \FOR{$t\leftarrow T_{i-1}+1$ to $T_i$}
                \STATE $\eta \leftarrow \frac{1}{2} [\sin(-\frac{\pi}{2} + \frac{t-T_{i-1}}{T_i-T_{i-1}}\pi) + 1]$
                \STATE $\lambda \leftarrow \lambda_{i-1} + \eta(\lambda_i-\lambda_{i-1})$
                \STATE update model by loss $\mathcal{L}_{lm}+\mathcal{L}_{reg}(\lambda)$
            \ENDFOR
        \ENDFOR
    \end{algorithmic}
\end{algorithm}
% \end{wrapfloat}

\section{Extented Introduction of Approximate Acceleration Algorithms} \label{sec:approximate-alg}

Existing approximate algorithms are mostly dependent on activation predictors, which are small neural networks to predict the intermediate activations $\mathbf{x}_1$ based on the input hidden states $\mathbf{x}$~\cite{liu2023deja,song2023powerinfer}. If one element at a specific position of $\mathbf{x}_1$ is predicted to be zero, then all the computations associated with this position can be saved with little or no hardware resources allocated. This is the key mechanism with which approximate algorithms realize acceleration.

Nevertheless, such a predictor-dependent acceleration effect is largely dependent on the performance of the pre-trained activation predictors. For example, a typical bad case is that an actually activated element in $\mathbf{x}_1$ is predicted to be inactivated. This can bring about negative results including unwise hardware resource allocation and erroneously ignored intermediate logits, which limits the practical speedup ratios and even causes inference inaccuracies. Therefore, a sparse LLM can gain more benefits from approximate algorithms if its activation distribution is more predictable.

To test a sparse LLM's practical acceleration value with approximate algorithms, we involve the predictability of its activation distribution, which is evaluated by the performance of its specifically pre-trained activation predictor. This involves two key metrics: the activation recall and the predicted sparsity.

The activation recall refers to the average ratio of correctly predicted activated elements among all the truly activated elements in $\mathbf{x}_1$. The predicted sparsity is calculated as the ratio of predicted inactivated elements among all the elements in $\mathbf{x}_1$. A predictor with higher recall will miss less truly activated elements, therefore reducing inference inaccuracies and bringing about wiser hardware allocation. Under comparable recalls, a higher predicted sparsity indicates fewer elements to be falsely predicted activated, which largely alleviates the waste of computational resources. These can help an acceleration framework obtain a better grasp of activation distribution and thus make wiser policies for faster inference as well as low inference inaccuracies~\cite{liu2023deja}.

% As for the evaluation metrics, the activation recall refers to the average ratio of correctly predicted activated elements among all the truly activated elements in $\mathbf{x}_1$. The predicted sparsity is calculated as the ratio of predicted inactivated elements among all the elements in $\mathbf{x}_1$. Predictors with higher recall and predicted sparsity can help an acceleration framework obtain a better grasp of activation distribution and thus make wiser policies for faster inference as well as low inference inaccuracies~\cite{liu2023deja}.

\section{Implementation Details of Sparse GPU Operators} \label{sec:impl-operator}

\textbf{Input-Side Sparse Operator.} The input-side sparse operator is a sparse matrix-vector multiplication operator for accelerating $\mathbf{x}_1 \mathbf{W}_2^T$, where the input $\mathbf{x}_1$ is sparse. 
Due to the sparsity of input, any operation involving a zero element in $\mathbf{x}_1$ can be omitted. 
Compared with a standard implementation of matrix-vector multiplication, both memory access and computation of the sparse operator will decrease with the sparsity increasing.

\textbf{Output-Side Sparse Operator.} The output-side sparse operator is a fused operator consisting of ReLU, sparse matrix-vector multiplication, and element-wise multiplication, for accelerating $\mathbf{s} \odot (\mathbf{x} \mathbf{W}_1^T)$, where $\mathbf{s}$ is sparse.
The sparsity of $\mathbf{s}$ can be propagated to the output of $\mathbf{x} \mathbf{W}_1^T$ through element-wise multiplication, which means that the computation of matrix-vector multiplication in $\mathbf{x} \mathbf{W}_1^T$ can be skipped whenever a result element will be multiplied by zero of sparse $\mathbf{s}$.
In addition, we postpone the ReLU activation function in $\sigma(\mathbf{x} \mathbf{W}_s^T)$ into this operator so that $\sigma$ can be implicitly performed along with the element-wise multiplication. 
These operations are fused into a single operator, thereby reducing the data movement between operations.

For implementation, we first load the result of $\mathbf{x} \mathbf{W}_s^T$, determine which elements are greater than zero (or a positive threshold after activation threshold shifting), and then select the corresponding columns of $\mathbf{W}_1^T$ to load from GPU memory, performing multiplication operations with $\mathbf{x}$. 
As the matrix $\mathbf{W}_1^T$ is sparse by column, we store the matrix in a column-major format to coalesce memory access and fully utilize vectorized loads/store instructions.
After this step, we get the sparse result vector of $\mathbf{x} \mathbf{W}_1^T$ and multiply the corresponding elements with activated elements of $\mathbf{s}$, with other elements filled with zeros directly.
Finally, the result vector $\mathbf{x}_1$ is obtained.

\section{Training and Evaluation Datasets} \label{sec:dataset-detail}

\paragraph{Training Datasets}

Our mixed training data consists of both language modeling datasets and instruction tuning datasets.
The language modeling datasets are directly cleaned and filtered from raw corpus, including StarCoder~\cite{li2023starcoder}, Wikipedia~\cite{wikidump}, Pile~\cite{gao2020pile}, and other collected datasets.
The instruction tuning datasets mainly involve input instructions and annotated target answers, including UltraChat~\cite{ding2023enhancing}, multiple-choice QA data of P3~\cite{sanh2021multitask} (Choice P3), PAQ~\cite{lewis2021paq}, Unnatural Instructions~\cite{honovich2022unnatural}, Flan~\cite{longpre2023flan}, Super-Natural Instructions~\cite{wang2022super}, and other collected datasets.

% Notably, the next-token prediction loss for these datasets is only computed on the target answers, excluding the input instructions.

\begin{table*}[ht]
    \centering
    \scriptsize
    \setlength{\tabcolsep}{1mm}
    \begin{tabular}{lcccccccccc}
        \toprule
        Setting & HumanEval & MBPP & PIQA & SIQA & HellaSwag & WinoGrande & COPA & BoolQ & LAMBADA & TyDi QA \\
        \cmidrule(lr){1-1}\cmidrule(lr){2-3}\cmidrule(lr){4-8}\cmidrule(lr){9-11}
        Original-7B & 10.98 & 21.77 & 78.40 & 47.70 & 75.67 & 67.17 & 79.00 & 75.99 & 72.81 & 36.82 \\
        ReluLLaMA-7B  & 12.20 & 19.51 & 77.86 & 49.54 & 72.85 & 64.96 & 83.00 & 78.10 & 70.33 & 63.18 \\
        % Vanilla ReLU-7B & 18.29 & 24.33 & 78.35 & 50.36 & 74.29 & 64.64 & 86.00 & 79.42 & 69.55 & 70.68 \\
        % Shifted ReLU-7B & 17.07 & 23.92 & 78.40 & 50.31 & 73.80 & 63.93 & 84.00 & 78.84 & 69.09 & 71.59 \\
        % Fixed $L_1$-7B & 17.07 & 20.64 & 76.06 & 44.32 & 68.30 & 63.38 & 78.00 & 52.08 & 65.01 & 49.09 \\
        \textbf{ProSparse-7B$^*$} & 16.46 & 22.48 & 75.79 & 43.50 & 71.08 & 64.09 & 77.00 & 62.48 & 67.73 & 59.77 \\
        \textbf{ProSparse-7B} & 16.46 & 22.38 & 75.68 & 43.55 & 71.09 & 64.01 & 77.00 & 62.51 & 68.21 & 59.77 \\
        \cmidrule(lr){1-1}\cmidrule(lr){2-3}\cmidrule(lr){4-8}\cmidrule(lr){9-11}
        Original-13B & 16.46 & 23.92 & 79.38 & 47.90 & 79.12 & 70.48 & 86.00 & 82.54 & 76.21 & 55.91 \\
        ReluLLaMA-13B & 17.07 & 23.31 & 78.40 & 47.13 & 76.60 & 69.06 & 81.00 & 81.16 & 73.49 & 65.23 \\
        \textbf{ProSparse-13B$^*$} & 25.61 & 32.44 & 77.04 & 45.14 & 75.91 & 68.67 & 82.00 & 79.27 & 71.08 & 52.27 \\
        \textbf{ProSparse-13B} & 23.78 & 33.06 & 77.26 & 45.29 & 75.88 & 68.35 & 82.00 & 78.93 & 71.36 & 50.45 \\
        \cmidrule(lr){1-1}\cmidrule(lr){2-3}\cmidrule(lr){4-8}\cmidrule(lr){9-11}
        MiniCPM-1B & 41.46 & 32.24 & 74.10 & 48.87 & 67.27 & 59.12 & 69.00 & 69.30 & 58.18 & 55.23 \\
        \textbf{ProSparse-1B$^*$} & 48.78 & 33.98 & 73.99 & 47.54 & 68.46 & 59.75 & 73.00 & 74.92 & 53.74 & 53.41 \\
        \textbf{ProSparse-1B} & 50.61 & 33.47 & 74.16 & 47.54 & 68.40 & 59.75 & 72.00 & 75.08 & 53.74 & 53.36 \\
        \bottomrule
    \end{tabular}
    \caption{The performance (\%) on each independent benchmark.}
    \label{tab:single-results}
\end{table*}

\begin{table*}[t]
    \centering
    \scriptsize
    \setlength{\tabcolsep}{1.2mm}
    \begin{threeparttable}
    \begin{tabular}{ccccccc}
        \toprule
        \multirow{2}{*}{Setting} & Accumulated & Average & Average & Accumulated & Average & Average \\
        & Tokens (B) & Sparsity (\%) & Performance (\%) & Tokens (B) & Sparsity (\%) & Performance (\%) \\
        \cmidrule(lr){1-1}\cmidrule(lr){2-4}\cmidrule(lr){5-7}
        % 673969_40000_fatrelu_0.01
        Vanilla ReLU & 34.60 & 66.04 & 41.40 & 89.13 & 64.93 & 41.52 \\
        Shifted ReLU & 34.60 & 69.59 & 41.33 & 89.13 & 68.35 & 41.40 \\
        % 673670_68000_fatrelu_0.01, 17000 * 8 * 2 * 4 * 32 * 1024 + (68000 - 17000) * 8 * 4 * 1 * 8 * 4096
        ProSparse    & 34.60 & 89.32 & 38.46 & 89.13 & 88.29 & 40.67 \\
        \bottomrule
    \end{tabular}
    \begin{tablenotes}
        $\dagger$ Note that ProSparse with 89.13B tokens applies different hyper-parameters (i.e., more training stages and step numbers) for a smoother trend of regularization factor increase and thus obtains higher performance than the 34.60B setting.
    \end{tablenotes}
    \captionsetup{skip=2mm}
    \caption{Comparison of ProSparse with two regularization-free former ReLUfication methods (7B). The bias $b$ for shifted ReLU is tuned to ensure the best performance.}
    \label{tab:baseline}
    \end{threeparttable}
\end{table*}

\paragraph{Evaluation Benchmarks}

To evaluate the task-specific performance of the LLMs obtained by ProSparse, we adopt the following comprehensive benchmarks.

(1) \textit{Code Generation}: We compute the average pass@1 scores on HumanEval (0-shot)~\cite{humaneval} and MBPP (3-shot)~\cite{mbpp}.
(2) \textit{Commonsense Reasoning}: We report the average 0-shot accuracies on PIQA~\cite{piqa}, SIQA~\cite{siqa}, HellaSwag~\cite{hellaswag}, WinoGrande~\cite{winogrande}, and COPA~\cite{copa}.
(3) \textit{Reading Comprehension}: We compute the average 0-shot accuracies on BoolQ~\cite{boolq}, LAMBADA~\cite{lambada}, and TyDi QA~\cite{tydiqa}.
(4) \textit{Other Popular Benchmarks}: We report the average accuracies on GSM8K (8-shot)~\cite{gsm8k}, MMLU (5-shot)~\cite{mmlu}, Big Bench Hard (BBH) (3-shot)~\cite{bbh}, and AGI-Eval (0-shot)~\cite{agieval}. Refer to Appendix~\ref{sec:eval-details} for more details.

\section{Training Details of Activation Predictors} \label{sec:train-act-predictor}

Following Deja Vu~\cite{liu2023deja}, the predictor is a two-layer FFN, composed of two linear projection layers with a ReLU activation in between them. Notably, as each layer of a sparse LLM has different activation distributions, we should introduce the same number of predictors as that of Transformer layers. For predictor training, we first collect about 400,000 pairs of input hidden states $\mathbf{x}$ and intermediate activations $\mathbf{x}_1$ at the corresponding layer. Next, we train the predictor on 95\% pairs with the binary cross entropy loss and compute the predictability metrics on the remaining 5\% pairs. We reserve the checkpoint with the highest recall to ensure the best inference accuracy with the least falsely ignored activations.

\section{Performance on Independent Benchmarks} \label{sec:performance-component}

In this section, we report the performance on each independent benchmark of Code Generation, Commonsense Reasoning, and Reading Comprehension, as displayed in Table~\ref{tab:single-results}.

\section{Performance Comparison between ProSparse and Baselines} \label{sec:comparison-baseline}

\begin{table*}[ht]
    \centering
    \scriptsize
    \setlength{\tabcolsep}{0.4mm}
    \begin{tabular}{cccccccc}
        \toprule
        Scale & $\lambda$ & Average Sparsity (\%) & Average Performance (\%) & Scale & $\lambda$ & Average Sparsity (\%) & Average Performance (\%) \\
        \cmidrule(lr){1-4}\cmidrule(lr){5-8}
        7B & $5e-2$ & 85.95 & 37.83    & 13B & $1e-2$ & 86.23 & 43.87 \\
        7B & $1e-1$ & 88.62 & 36.34    & 13B & $2e-2$ & 88.96 & 42.85 \\
        7B & $5e-1$ & 93.15 & 33.86    & 13B & $5e-2$ & 92.65 & 40.13 \\
        \cmidrule(lr){1-4}\cmidrule(lr){5-8}
        7B & ProSparse & 89.32 & \textbf{38.46} & 13B & ProSparse & 88.80 & \textbf{44.90} \\
        \bottomrule
    \end{tabular}
    \caption{Ablation study results about progressive sparsity regularization, the second step of ProSparse. $\lambda$ refers to the constant regularization factor in the second stage of ablation settings.}
    \label{tab:ablation-progress}
\end{table*}

Given the different amounts of training tokens, we compare ProSparse with the other two baselines (i.e., vanilla ReLU and shifted ReLU) in terms of the average sparsity and performance. As shown in Table~\ref{tab:baseline}, ProSparse can achieve far higher sparsity than two baselines. Besides, with more training tokens given, ProSparse is able to apply a smoother trend of regularization increase and thus better mitigate performance degradation. This is why the performance gap between ProSparse and two regularization-free baselines is bridged when the tokens increase from 34.60B to 89.13B. The additional training costs, namely 54.53B tokens, only account for about 2.73\% of the pre-training costs of the original LLaMA2~\cite{touvron2023llama2} and are well acceptable.

\section{Ablation Studies of Progressive Sparsity Regularization} \label{sec:ablation-progress}

Here we provide more detailed experimental results about the ablation of progressive sparsity regularization, as shown in Table~\ref{tab:ablation-progress}. Note that for ablation settings, we keep the regularization factor constant without progressive increase. More specifically, the whole training process consists of three steps: activation function substitution, continual training with a constant regularization factor, and activation threshold shifting.

\section{SFT for Sparsely Activated Models} \label{sec:sft-sparse}

The key problem for SFT sparsely activated models is how to feed instruction following knowledge into the model while maintaining the sparsity simultaneously. From the above observations about training dynamics, the regularization factor is still indispensable during SFT to avoid a considerable drop in sparsity. Moreover, the factor is probably not equal to the one used in the final progressive regularization stage (i.e., $\lambda_S$), as the data distribution has shifted radically.

Our practice of training ProSparse-1B can provide empirical answers to this problem. Concretely, while ProSparse-7B and ProSparse-13B are directly trained from the original LLaMA2 on mixed data through the three-step ProSparse, the training of ProSparse-1B includes an extra decay stage and an SFT stage, following the original practice of MiniCPM for better performance (see Table~\ref{tab:hyper-param-1b}). The decay stage is conducted on the mixed data, with a decreasing learning rate and a fixed regularization factor of value $\lambda_S$. By contrast, the SFT stage is performed only on the instruction tuning data. We find that \textbf{the regularization factor for SFT should be empirically smaller than $\lambda_S$ in order to accommodate newly injected knowledge from SFT data and avoid performance degradation}. For ProSparse-1B, an SFT factor of $1e-2$ works the best with an average performance of 44.72\%, while the performance drops to 44.32\% with $\lambda_S=5e-2$. Therefore, \textbf{SFT can be applied to sparsely activated models obtained by ProSparse with a well-chosen regularization factor}.

\section{Evaluation Details} \label{sec:eval-details}

For evaluation benchmarks including PIQA, SIQA, HellaSwag, WinoGrande, COPA, BoolQ, LAMBADA, TyDi QA, and AGI-Eval, we obtain the predicted answers based on maximized perplexity. Specifically, the predicted answer to a given question corresponds to the candidate that produces the lowest perplexity when it is concatenated to the question. For GSM8K, MMLU, and BBH, the predicted answers are determined by the option numbers directly generated by the models.

\begin{table*}[ht]
    \centering
    \scriptsize
    \setlength{\tabcolsep}{1mm}
    \begin{tabular}{cccccccc}
        \toprule
        \multicolumn{4}{c}{ProSparse-7B} & \multicolumn{4}{c}{ProSparse-13B} \\
        \cmidrule(lr){1-4}\cmidrule(lr){5-8}
        Stage Number $i$ & $\lambda_i$ & $T_i$ & Accumulated Tokens (B) & Stage Number $i$ & $\lambda_i$ & $T_i$ & Accumulated Tokens (B) \\
        \cmidrule(lr){1-4}\cmidrule(lr){5-8}
        0 & 0 & 5,000 & 10.49 & 0 & 0 & 5,500 & 46.14 \\
        1 & $5e-3$ & 6,000 & 12.58 & 1 & $5e-3$ & 6,750 & 56.62 \\
        2 & $5e-2$ & 10,000 & 20.97 & 2 & $1e-2$ & 10,750 & 90.18 \\
        3 & $5e-2$ & 12,000 & 25.17 & 3 & $1e-2$ & 11,000  & 92.27 \\
        4 & $2e-1$ & 16,000 & 33.55 & 4 & $2e-2$ & 15,000 & 125.83 \\
        5 & $2e-1$ & 16,500  & 34.60 & 5 & $2e-2$ & 16,000 & 134.22 \\
        \bottomrule
    \end{tabular}
    \caption{The important hyperparameters for training ProSparse-7B and ProSparse-13B. For simplicity, the 0th stage refers to the continual training in activation function substitution. The 1st stage is the warmup stage with a fixed regularization factor $\lambda_1$. The remaining stages are incremental stages with an increasing factor.}
    \label{tab:hyper-param}
\end{table*}

\section{Important Hyperparameters} \label{sec:hyper-param}

\begin{table}[ht]
    \centering
    \scriptsize
    \setlength{\tabcolsep}{0.5mm}
    \begin{tabular}{cccc}
        \toprule
        \multicolumn{4}{c}{ProSparse-1B} \\
        \cmidrule(lr){1-4}
        Stage Number $i$ & $\lambda_i$ & $T_i$ & Accumulated Tokens (B) \\
        \cmidrule(lr){1-4}
        0 & 0 & 10,000 & 49.15 \\  % 325000 -> 335000
        1 & $1e-3$ & 15,000 & 73.73 \\  % 335000 -> 340000
        2 & $5e-3$ & 20,000 & 98.30 \\  % 340000 -> 350000
        3 & $5e-3$ & 25,000 & 122.88 \\  % 340000 -> 350000
        4 & $5e-2$ & 35,000 & 172.03 \\  % 350000 -> 360000
        \cmidrule(lr){1-4}
        decay & $5e-2$ (fixed) & 95,000 & 466.94 \\  % 360000 -> 420000
        SFT   & $1e-2$ (fixed) & 101,000  & 473.02 \\  % 0 -> 6000
        \bottomrule
    \end{tabular}
    \caption{The important hyperparameters for ProSparse-1B. Compared with the other two settings, we follow the original practice of MiniCPM-1B~\cite{hu2024minicpm}, appending an extra decay stage and an SFT stage. Note that each of the additional stages has a constant regularization factor.}
    \label{tab:hyper-param-1b}
\end{table}

We provide the important hyperparameters for ProSparse training, as shown in Table~\ref{tab:hyper-param} and Table~\ref{tab:hyper-param-1b}. Note that the peak regularization factors of two contiguous stages can be set to the same value to introduce an extra constant-factor stage, mainly for stability requirements. For ProSparse-7B and ProSparse-13B, We use a cosine annealing learning rate scheduler throughout the training process and the peak learning rates are $3e-5$ and $5e-5$ for 7B and 13B respectively. For ProSparse-1B, we use exactly the same hyper-parameter settings as MiniCPM-1B~\cite{hu2024minicpm} except for the $L_1$ regularization. After pre-training on the language modeling dataset with the paradigm of ProSparse, following the original practice, we add an extra decay stage (mixed data with a decreasing learning rate) and an SFT stage (only instruction tuning data with a fixed learning rate). Each of the additional stages has a constant regularization factor. The context length is 4,096 for all settings. Considering cost issues, the hyper-parameters for ProSparse are set to appropriate values to just match the original Swish-activated versions in terms of benchmark performance.

% All the baseline models are trained with the same number of tokens and the same mixed training dataset as ProSparse.

All the 7B models are trained with the AdamW optimizer on 8 A100 80GB GPUs for about 10 days. All the 13B models are trained on 32 A100 80GB GPUs for about 20-30 days. The LLMs of each method involved in this paper are trained once due to the formidable training costs.

% \newpage

\section{Dataset-Wise Sparsity Distribution} \label{sec:dataset-dis}

\begin{table*}[ht]
    \centering
    \scriptsize
    \setlength{\tabcolsep}{1.0mm}
    \begin{tabular}{lcccccccccc}
        \toprule
        \multirow{2}{*}{Setting} & \multirow{2}{*}{Mixed} & \multirow{2}{*}{StarCoder} & \multirow{2}{*}{Wikipedia} & \multirow{2}{*}{Pile} & \multirow{2}{*}{UltraChat} & Choice & \multirow{2}{*}{PAQ} & \multirow{2}{*}{Flan} & Unnatural & Super-Natural \\
        & & & & & & P3 & & & Instructions & Instructions \\
        \cmidrule(lr){1-1}\cmidrule(lr){2-2}\cmidrule(lr){3-5}\cmidrule(lr){6-11}
        ReluLLaMA-7B  & 66.98 & 66.60 & 67.16 & 67.35 & 67.91 & 67.35 & 66.98 & 67.35 & 66.42 & 66.98 \\
        % Vanilla ReLU-7B & 66.04 & 65.86 & 65.67 & 65.86 & 67.16 & 66.42 & 66.23 & 65.86 & 65.49 & 65.86 \\
        % Shifted ReLU-7B & 69.59 & 69.59 & 69.03 & 69.03 & 70.52 & 69.78 & 69.40 & 69.22 & 69.22 & 69.03 \\
        % Fixed $L_1$-7B & 91.46 & 91.23 & 87.97 & 87.97 & 95.45 & 99.33 & 98.58 & 93.52 & 96.20 & 98.01 \\
        \textbf{ProSparse-7B$^*$}  & 88.11 & 88.20 & 83.30 & 84.24 & 91.23 & 97.94 & 96.74 & 90.76 & 93.00 & 95.71 \\
        \textbf{ProSparse-7B} & 89.32 & 89.13 & 84.33 & 85.35 & 93.66 & 98.33 & 97.28 & 91.74 & 93.80 & 96.32 \\
        \cmidrule(lr){1-1}\cmidrule(lr){2-2}\cmidrule(lr){3-5}\cmidrule(lr){6-11}
        ReluLLaMA-13B & 71.56 & 71.33 & 71.45 & 71.56 & 72.27 & 71.80 & 71.21 & 71.56 & 70.85 & 71.33 \\
        \textbf{ProSparse-13B$^*$}  & 87.97 & 87.50 & 81.64 & 83.06 & 92.45 & 98.41 & 97.54 & 91.65 & 92.92 & 96.40 \\
        \textbf{ProSparse-13B} & 88.80 & 88.63 & 83.65 & 84.12 & 92.65 & 98.73 & 97.99 & 92.54 & 93.66 & 96.92 \\
        \cmidrule(lr){1-1}\cmidrule(lr){2-2}\cmidrule(lr){3-5}\cmidrule(lr){6-11}
        \textbf{ProSparse-1B$^*$} & 86.25 & 86.84 & 82.72 & 83.23 & 88.50 & 89.83 & 83.36 & 83.93 & 90.16 & 90.70 \\
        \textbf{ProSparse-1B} & 87.89 & 88.44 & 84.71 & 85.17 & 89.93 & 91.01 & 85.36 & 85.82 & 91.36 & 91.76 \\
        \bottomrule
    \end{tabular}
    \caption{The average sparsity (\%) on our mixed training dataset (denoted as ``Mixed'') and its components, divided into language modeling datasets and instruction tuning datasets.}
    \label{tab:dataset-sparsity}
\end{table*}

Despite the satisfactory average sparsity, there still exist gaps between the mixed training dataset and the actual input texts that the model will encounter in real-life applications. To investigate the sparsity of our model under different scenarios, we compute the sparsity on each component of the mixed training data respectively.

As demonstrated in Table~\ref{tab:dataset-sparsity}, the sparse LLMs obtained through ProSparse have a pronounced property of inconsistent dataset-wise sparsity. Concretely, the sparsity on instruction tuning datasets is significantly higher than those on language modeling datasets (i.e., StarCoder, Wikipedia, and Pile). Considering the contents of datasets, we come to the following assumption: \textbf{the more formatted a dataset is, 
the less hybrid knowledge is needed for generation, and thus the $L_1$-regularized models can achieve higher activation sparsity with fewer neurons activated.} Plain text datasets including Wikipedia and Pile have the lowest sparsity, followed by the more formatted code dataset StarCoder. Among instruction tuning datasets, QA datasets (e.g., Choice P3) with the most monotonic input-output formats obtain the highest sparsity. By contrast, the sparsity is relatively lower on UltraChat and Flan, covering general dialogues and a wide variety of tasks respectively. Notably, dialogues and tasks with formatted instructions cover a majority of input contents of conversational AI, the mainstream application form of LLMs. Such higher sparsity on instruction tuning data will endow ProSparse with more significant practical values.

\begin{table*}[ht]
    \centering
    \scriptsize
    \setlength{\tabcolsep}{0.6mm}
    \begin{tabular}{lccccccccc}
        \toprule
        \multirow{2}{*}{Setting} & Average & Code & Commonsense & Reading & \multirow{2}{*}{GSM8K} & \multirow{2}{*}{MMLU} & \multirow{2}{*}{BBH} & \multirow{2}{*}{AGI Eval} & Average \\
        & Sparsity & Generation & Reasoning & Comprehension & & & & & Performance \\
        \cmidrule(lr){1-1}\cmidrule(lr){2-2}\cmidrule(lr){3-9}\cmidrule(lr){10-10}
        ProSparse-7B$^*$ & 88.11 & 19.47 & 66.29 & 63.33 & 12.74 & 45.21 & 33.59 & 27.55 & 38.31 \\
        ProSparse-7B $t=0.005$ & 88.62 & 19.68 & 66.23 & 62.59 & 12.05 & 44.95 & 34.43 & 27.46 & 38.20 \\
        ProSparse-7B $t=0.01$  & 89.32 & 19.42 & 66.27 & 63.50 & 12.13 & 45.48 & 34.99 & 27.46 & 38.46 \\
        ProSparse-7B $t=0.02$  & 90.35 & 18.39 & 66.09 & 62.93 & 12.59 & 45.02 & 34.34 & 27.14 & 38.07 \\
        ProSparse-7B $t=0.03$  & 90.95 & 18.65 & 66.24 & 62.72 & 12.13 & 44.83 & 34.92 & 27.36 & 38.12 \\
        \cmidrule(lr){1-1}\cmidrule(lr){2-2}\cmidrule(lr){3-9}\cmidrule(lr){10-10}
        ProSparse-13B$^*$       & 87.97 & 29.03 & 69.75 & 67.54 & 25.40 & 54.78 & 40.20 & 28.76 & 45.07 \\
        ProSparse-13B $t=0.005$ & 88.24 & 29.04 & 69.69 & 67.62 & 26.23 & 54.75 & 39.52 & 28.74 & 45.08 \\
        ProSparse-13B $t=0.01$  & 88.80 & 28.42 & 69.76 & 66.91 & 26.31 & 54.35 & 39.90 & 28.67 & 44.90 \\
        ProSparse-13B $t=0.02$  & 89.40 & 29.29 & 69.63 & 65.28 & 24.94 & 54.88 & 39.79 & 28.88 & 44.67 \\
        ProSparse-13B $t=0.03$  & 90.23 & 28.12 & 69.28 & 64.79 & 25.85 & 54.68 & 40.08 & 28.71 & 44.50 \\
        \bottomrule
    \end{tabular}
    \caption{The sparsity (\%) and performance (\%) under different thresholds $t$ of the activation threshold shifting step.}
    \label{tab:different-thred}
\end{table*}

\section{Layer-Wise Sparsity Distribution} \label{sec:layer-dis}

\begin{figure}[ht]
    \centering
    \includegraphics[width=1.0\linewidth]{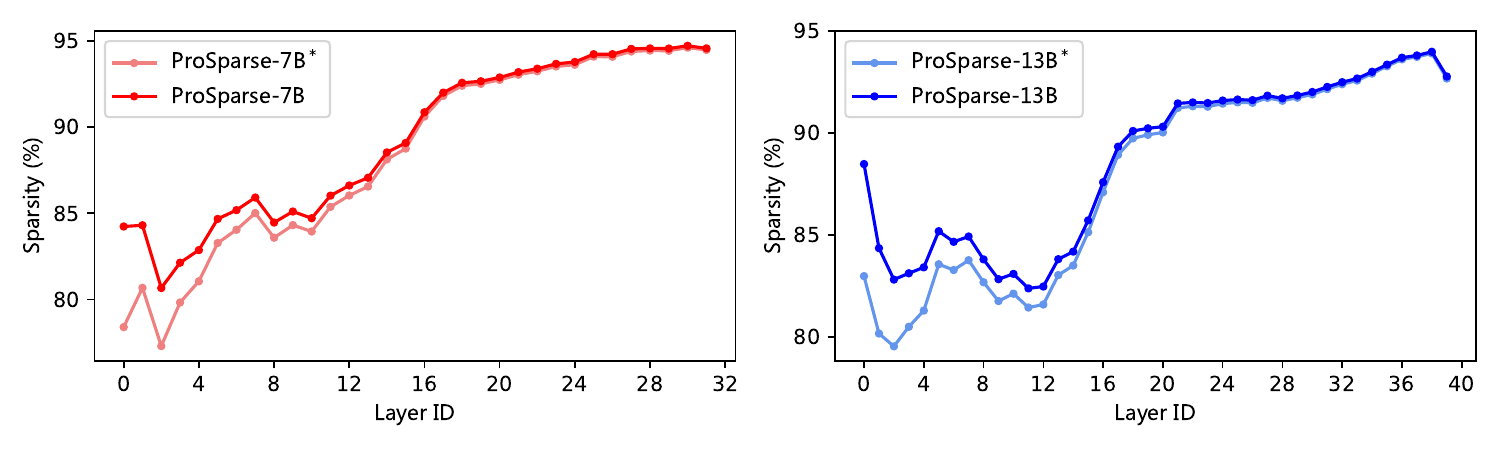}
    \caption{The layer-wise sparsity of ProSparse models. The marker ``$^*$'' denotes the settings without activation threshold shifting.}
    \label{fig:layer-curve}
    \vspace{2mm}
\end{figure}

Another problem worth concern is the layer-wise sparsity, which potentially impacts the load balance and the inference efficiency. Therefore, we compute the sparsity of each layer for ProSparse models, as shown in Figure~\ref{fig:layer-curve}.

From the tendency of the line chart, we clearly observe layer-wise sparsity imbalance in that lower layers are significantly denser than higher layers. Nevertheless, the activation threshold shifting can considerably improve the sparsity of lower layers with little impact on higher layers. Although this technique only contributes marginally to the average sparsity, it is still indispensable in alleviating the layer-wise sparsity imbalance issue.

\section{Effect of Different Thresholds in Activation Threshold Shifting} \label{sec:different-thres}

As mentioned in Section~\ref{sec:thres-adj}, the threshold $t$ is an important hyper-parameter in activation threshold shifting, the last step of ProSparse. In the overall experimental results, we choose $t=0.01$ for both ProSparse-7B and ProSparse-13B to balance the sparsity and performance. Here we list the results under other thresholds in Table~\ref{tab:different-thred}. As can be observed, a small $t$ results in a quite limited sparsity improvement compared with the version without activation threshold shifting, while a larger $t$ can cause more performance degradation. Therefore, we choose $t=0.01$ to strike a balance.

\end{document}